\documentclass[10pt,twocolumn]{article}

\usepackage[margin=0.75in]{geometry}
\usepackage[T1]{fontenc}
\usepackage{lmodern}
\usepackage{microtype}
\usepackage{amsmath}
\usepackage{amssymb}
\usepackage{booktabs}
\usepackage{graphicx}
\usepackage{xcolor}
\usepackage{tikz}
\usepackage{url}
\usepackage{hyperref}
\usetikzlibrary{arrows.meta,positioning,shapes.geometric,fit}

\hypersetup{
 colorlinks=true,
 linkcolor=blue!50!black,
 citecolor=blue!50!black,
 urlcolor=blue!50!black,
 pdftitle={The Constraint Tax: Measuring Validity-Correctness Tradeoffs in Structured Outputs for Small Language Models},
 pdfauthor={Jaideep Ray}
}

\definecolor{ctblue}{HTML}{DDEBFF}
\definecolor{ctgreen}{HTML}{DFF5E1}
\definecolor{ctorange}{HTML}{FFE9C9}
\definecolor{ctred}{HTML}{FFD6D6}
\definecolor{ctpurple}{HTML}{E9DDFF}
\definecolor{ctgray}{HTML}{F2F2F2}

\title{The Constraint Tax: Measuring Validity-Correctness Tradeoffs in Structured Outputs for Small Language Models}
\author{Jaideep Ray\\\texttt{jaray@acm.org}}
\date{}

\newcommand{\tax}{\mathrm{Tax}}
\newcommand{\acc}{\mathrm{Acc}}

\newcommand{\mode}[1]{\texttt{#1}}

\begin{document}
\sloppy
\maketitle

\begin{abstract}
Many deployed LLM applications pass model completions directly into software: JSON envelopes, typed traces, regex-limited fields, and tool-call arguments. We study that interface in on-device and low-cost small language model (SLM) deployments, where sub-3B checkpoints are attractive for privacy, latency, and commodity hardware but have little spare capacity for solving and formatting at the same time. The usual assumption is that hard output constraints only improve the wrapper around an answer. Our measurements show that, for small models, the wrapper can change the answer. We define \emph{constraint tax}, a paired protocol for measuring answer and executable-accuracy loss caused by structured-output constraints at fixed model, task distribution, and problem instances. We report five result settings: the main commodity-GPU suite, a calendar tool-call analogue, a 3B boundary check, backend replication, and a local expanded-interface study. In the 15,000-generation main suite, hard answer-only schema decoding raises schema validity from 61.5\% to 100.0\%, but lowers answer accuracy from 19.7\% to 11.0\% and increases wrong-valid-schema outputs from 49.5\% to 88.9\%. In the calendar analogue, prompt-only JSON achieves 91.5\% executable accuracy; the same hard tool-call schema reaches only 48.0\%, even though both directly generated modes are 100.0\% schema-valid. The error is semantic, not structural. We also find a direct-schema tax at the 3B boundary and use delayed packaging as evidence for a practical pattern: reason free, constrain late. Production reports should separate schema validity, answer accuracy, executable accuracy, and wrong-valid-schema rate.
\end{abstract}

\section{Introduction}

Structured outputs are now a serving contract. Agents pass JSON to tools, gateways validate schemas, routers expect typed fields, and product code often treats parseability as the gate before execution. That contract is useful, but it is not neutral. A sub-3B local model has less decoding capacity and weaker instruction following than a large hosted model, so the same schema can compete with the task it is meant to package.

The target setting is on-device SLM deployment: private assistants, offline app agents, local enterprise workflows, and edge tools that need executable outputs without routing every request to a hosted model. These systems still need schemas for tool calls, UI actions, and local APIs. They also operate under tighter memory, latency, and model-capacity budgets. Under those constraints, reasoning and output control become part of the same systems problem.

Structured output is not merely post-processing. During constrained decoding it changes which tokens are legal, which intermediate representations can appear, and how much of the completion budget is spent on syntax rather than problem solving. A valid JSON object can still encode the wrong decision, so a dashboard that tracks parse success alone can improve while downstream execution gets worse.

We treat this as a measurement problem rather than a decoder-design problem. The question is: when a small model is forced into a schema, how much semantic correctness is lost relative to comparable unconstrained or prompt-only baselines?

The paper contributes the following pieces.

\begin{enumerate}
 \item We define \emph{constraint tax}, an accuracy delta between a baseline interface and a structured-output mode under the same model and task distribution.
 \item We separate the metrics that are often collapsed in deployment logs: schema validity, answer accuracy, executable accuracy, trace correctness, latency, token count, structural overhead, and wrong-valid-schema rate.
 \item We provide a low-compute harness that runs MLX inference for unconstrained modes and vLLM/SGLang constrained decoding on commodity GPUs.
 \item We measure a silent executable-correctness regression: in a calendar tool-call analogue, both interfaces are 100.0\% schema-valid, but hard schema decoding loses 43.5 points of executable accuracy.
 \item We identify the constructive design pattern \emph{reason free, constrain late}: use the least intrusive constraint that satisfies the downstream contract, and package answers only after the model has solved the task.
\end{enumerate}

\section{Background and Motivation}

\subsection{Output Constraints as a Reasoning Intervention}

Constrained decoding masks tokens that would violate a grammar, regular expression, or JSON schema. That is appropriate when syntax is the main objective. It is less benign when the model needs a scratchpad. A small model may have to externalize partial arithmetic, symbolic state, or intermediate decisions. A rigid schema can push those states through field names, quoted strings, arrays, object delimiters, and type constraints that were chosen for software consumption rather than reasoning.

The interaction can be summarized as:

\begin{equation}
Q = f(R_{\mathrm{reason}}, R_{\mathrm{format}}, B_{\mathrm{decode}}),
\end{equation}

where quality depends on reasoning resources, formatting resources, and decoding budget. For a large model, format fidelity may consume a small fraction of effective capacity. For a sub-3B model, the same schema can be a material part of the generation problem. The concerning case is not merely invalid JSON; it is \emph{wrong answer, valid schema}.

\subsection{Structured Decoding as a Serving-System Interface}

The failure mode crosses the model-serving boundary. A product system may use one backend for freeform generation, another for grammar-constrained decoding, and a third layer for parsing and validation. The measured behavior is a property of the model, prompt, decoding algorithm, schema, and validation stack together. Model accuracy alone or parser success alone misses the interaction.

For that reason, we treat the structured-output path as a full system. Each record stores the prompt, schema or regex, backend, raw generation, parse outcome, executable checker result, trace checker result, latency, and output-token count. This supports comparisons across prompt-only JSON, regex-constrained answers, direct JSON schema modes, vLLM, SGLang, and MLX execution under a common logging format.

\section{Measurement Design}

\begin{figure*}[t]
\centering
\resizebox{\textwidth}{!}{%
\begin{tikzpicture}[
 font=\small,
 node distance=0.75cm and 0.95cm,
 box/.style={draw=black!65, very thick, rounded corners=3pt, align=center, minimum height=0.82cm, inner sep=5pt},
 arrow/.style={-{Latex[length=2.2mm]}, very thick, draw=black!70},
 dashedbox/.style={draw=black!55, thick, rounded corners=4pt, dashed, inner sep=6pt}
]

\node[box, fill=ctblue, minimum width=2.4cm] (tasks) {Deterministic\\task generator};
\node[box, fill=ctorange, minimum width=2.4cm, right=of tasks] (modes) {Prompt + mode\\freeform, regex, schema};
\node[box, fill=ctpurple, minimum width=2.35cm, right=of modes] (router) {Backend router\\MLX, HF, vLLM, SGLang};
\node[box, fill=ctgreen, minimum width=2.55cm, right=of router] (decode) {Generation\\constrained or free};
\node[box, fill=ctred, minimum width=2.55cm, right=of decode] (eval) {Parser + validator\\executor + trace check};
\node[box, fill=ctgray, minimum width=2.45cm, right=of eval] (metrics) {Records\\tax, validity, latency};

\draw[arrow] (tasks) -- (modes);
\draw[arrow] (modes) -- (router);
\draw[arrow] (router) -- (decode);
\draw[arrow] (decode) -- (eval);
\draw[arrow] (eval) -- (metrics);

\node[dashedbox, fit=(modes) (router) (decode), label={[font=\small]above:Output-control intervention}] {};
\node[dashedbox, fit=(eval) (metrics), label={[font=\small]below:Measurement layer}] {};

\end{tikzpicture}%
}
\caption{Constraint-tax harness. The benchmark keeps problem instances fixed while varying the output-control intervention. The measurement layer records syntax outcomes and semantic correctness separately.}
\label{fig:system}
\end{figure*}
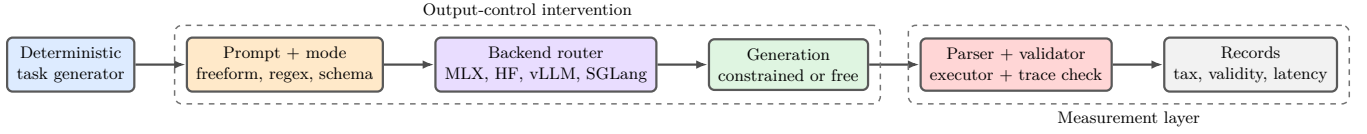

Figure~\ref{fig:system} summarizes the harness used throughout the experiments.

\subsection{Tasks}

The benchmark uses deterministic synthetic tasks with exact answer normalization. Synthetic tasks are deliberate. They do not claim to cover broad user utility, but they provide controlled ground truth, reproducible task generation, and executable answer checking. This makes them appropriate for isolating the effect of output constraints without using another language model as a judge.

The benchmark reports five result settings. The main run covers five deterministic task families across a small-model suite. A separate production analogue uses an executable calendar-object task in which outputs are checked as tool-call arguments. For tasks that expose intermediate reasoning, the harness also supports trace checking. The trace checker verifies expected step count and normalized final intermediate outputs. Table~\ref{tab:taskfamilies} lists the generator-backed family names, prompt templates, and ground-truth formats.

\begin{table*}[t]
\centering
\caption{StructReason-Small task families as implemented by the deterministic generator. Ground truth is stored per instance as a final answer string plus typed trace steps.}
\label{tab:taskfamilies}
\resizebox{\textwidth}{!}{%
\begin{tabular}{lll}
\toprule
Family & Prompt template & Ground-truth format \\
\midrule
\mode{arithmetic\_two\_step} & A box has $r$ red balls and $b$ blue balls. Sam adds $a$ red balls and removes $m$ blue balls. How many balls are left? & Integer answer; trace operations \mode{initial\_total}, \mode{add\_red}, \mode{remove\_blue}, \mode{final\_total}. \\
\mode{symbolic\_string} & Take the last letter of each word: $w_1,\ldots,w_k$. Concatenate them. & Concatenated last-letter string; one \mode{last\_letter} trace step per word plus \mode{concatenate}. \\
\mode{object\_tracking} & $p_1$ has the key. $p_2$ has the $i_2$. $p_3$ has the $i_3$. $p_1$ and $p_2$ swap items. $p_2$ and $p_3$ swap items. Who has the key? & Holder name; trace operations \mode{initial\_state}, two \mode{swap} steps, and \mode{holder\_of\_key}. \\
\mode{boolean\_logic} & A is true/false. B is true/false. C = A AND NOT B. D = C OR B. What is D? & Boolean string \mode{true} or \mode{false}; trace operations \mode{not}, \mode{and}, \mode{or}. \\
\mode{tool\_call\_argument} & Assume today is $d$. User request: Schedule a $u$ minute meeting with attendee relative-date at display-time about topic. Available tools: \mode{create\_calendar\_event}(\ldots). Return the selected tool and arguments. Do not access a real calendar. & Canonical JSON object with \mode{tool} and sorted \mode{arguments}; trace operations \mode{select\_tool}, \mode{resolve\_date}, \mode{resolve\_time}, \mode{build\_arguments}. \\
\bottomrule
\end{tabular}%
}
\end{table*}

\subsection{Output Modes}

The harness compares interfaces that differ in how much structure is imposed during reasoning; Table~\ref{tab:modes} lists the evaluated modes.

\begin{table}[t]
\centering
\caption{Output modes used by the benchmark. The modes separate prompt-only formatting, regex constraints, direct schemas, structured rationales, typed traces, and delayed packaging.}
\label{tab:modes}
\resizebox{\linewidth}{!}{%
\begin{tabular}{ll}
\toprule
Mode & Interface type \\
\midrule
\mode{freeform} & Verbose step-by-step response \\
\mode{freeform\_direct} & Answer-only final line \\
\mode{freeform\_brief\_reasoning} & Short scratchpad, final answer last \\
\mode{prompt\_json} & JSON requested only by prompt \\
\mode{final\_only\_regex} & Regex-constrained final answer \\
\mode{answer\_only\_schema} & Minimal answer JSON schema \\
\mode{rationale\_answer\_schema} & JSON schema with rationale and answer \\
\mode{typed\_trace\_schema} & Typed JSON trace with final answer \\
\mode{delayed\_constraint} & Reason first, package answer second \\
\bottomrule
\end{tabular}%
}
\end{table}

\mode{prompt\_json} measures whether natural-language formatting instructions are enough. \mode{final\_only\_regex} measures whether a narrow final-answer constraint avoids most reasoning interference. Direct schema modes measure hard grammar pressure. \mode{typed\_trace\_schema} tests whether making reasoning itself structured helps or hurts small models. \mode{delayed\_constraint} tests whether structured packaging should happen after the model has solved the problem.

\subsection{Backends}

The benchmark is designed for low-cost execution. Mac-local experiments use MLX or Hugging Face Transformers, depending on model support. Constrained decoding is routed to commodity GPU runs using vLLM or SGLang OpenAI-compatible servers. A 3B checkpoint is used as a boundary check rather than a default sub-3B model.

This split creates an interpretation rule: latency comparisons are most meaningful within backend families unless the deployment environment is normalized. Accuracy and validity comparisons are still replicated across decoding engines where possible.

\section{Metrics}

\subsection{Syntax and Semantic Metrics}

For each generation, the harness records:

\begin{itemize}
 \item \textbf{Schema validity:} whether the output parses and satisfies the required schema or regex.
 \item \textbf{Answer correctness:} whether the normalized answer matches deterministic ground truth.
 \item \textbf{Executable accuracy:} whether a task-specific checker accepts the produced object or answer.
 \item \textbf{Trace correctness:} whether expected trace properties match.
 \item \textbf{Wrong-valid-schema rate:} the fraction of examples that are schema-valid but fail the task-relevant answer or executable checker.
 \item \textbf{Structural overhead, latency, and tokens:} serving and interface-cost indicators.
\end{itemize}

These metrics intentionally distinguish validity from correctness. A production parser only sees whether the object is consumable. The benchmark additionally checks whether the consumable object is right.

\subsection{Constraint Tax}

Let $m$ be a model, $t$ a task family, $c$ a constrained output mode, and $b$ a baseline interface for the same problem instances. We define absolute constraint tax as:

\begin{equation}
\tax(m,t,c;b) = \max\left(0, \acc(m,t,b) - \acc(m,t,c)\right).
\end{equation}

The normalized form is:

\begin{equation}
\tax_{norm}(m,t,c;b) = \frac{\tax(m,t,c;b)}{\max(\epsilon, \acc(m,t,b))}.
\end{equation}

Here $\acc$ is the task-relevant semantic metric, either answer accuracy or executable accuracy. The clipping at zero is intentional. The metric asks how much accuracy is lost relative to a baseline. If a constraint improves accuracy, it is reported as an accuracy gain, not as negative tax. When available, we report 95\% bootstrap confidence intervals over problem instances.

\subsection{Statistical Reporting}

For paired comparisons, deltas are computed on the same generated problem instances whenever the experiment design supports pairing. Confidence intervals reported in the tables are nonparametric bootstrap intervals over problem instances. We report intervals only for runs where they are present in the generated result artifacts. For that reason, the MLX expanded-interface study is design evidence rather than the primary significance claim.

\subsection{Error Taxonomy}

Outputs are assigned to interpretable error classes: \mode{correct\_valid}, \mode{invalid\_json}, \mode{parse\_failure\_freeform}, \mode{schema\_validation\_error}, \mode{trace\_answer\_contradiction}, or \mode{wrong\_answer\_valid\_schema}. This taxonomy is more actionable than a pass/fail bit. Parser retries help invalid JSON; they do not fix wrong valid answers.

\section{Experimental Protocol}

All reported numbers come from generated JSONL or CSV artifacts in the experiment directories. We do not infer numbers from model expectations or hand-labeled outputs.

\textbf{Main GPU run.} We run vLLM on a commodity GPU with the main small-model suite. Each checkpoint is evaluated on 1,000 problem instances across five deterministic task families under each of five modes: \mode{freeform}, \mode{freeform\_direct}, \mode{freeform\_brief\_reasoning}, \mode{prompt\_json}, and \mode{answer\_only\_schema}. This yields 15,000 generations.

\textbf{Production analogue.} We run one main-suite checkpoint on a deterministic calendar tool-call argument task. The model must produce an object with tool name and arguments: title, date, start time, duration, attendee, and topic. We compare prompt-only JSON against the same required object enforced as a hard schema over 200 examples per directly generated mode, and include a derived delayed-packaging control that deterministically re-serializes the prompt-only JSON first-stage records.

\textbf{Backend replication and 3B boundary.} We replicate \mode{answer\_only\_schema} across vLLM and SGLang for the main-suite checkpoints. We also run the 3B boundary checkpoint with \mode{prompt\_json} and \mode{answer\_only\_schema} to test whether the direct-schema tax disappears near the 3B boundary.

\textbf{Local expanded-interface study.} We run an MLX-local study over four public small models, 20 examples per task family, and the full interface set, including rationale schemas, typed traces, delayed constraints, and regex modes. This study has fewer examples per task family than the main direct-comparison study, so we use it primarily for interface-design insight.

\section{Empirical Results}

The results start with the broad main-run aggregate and then move to the calendar tool-call analogue. The calendar task is the sharper test: validity is already saturated, so the execution drop cannot be explained away as a parse failure.

\subsection{Validity Improves While Correctness Falls}

\begin{table}[t]
\centering
\caption{Main-suite aggregate over the small-model checkpoints. Hard schema decoding buys validity but shifts errors into wrong-valid-schema outputs. Brackets denote 95\% bootstrap confidence intervals.}
\label{tab:main}
\resizebox{\linewidth}{!}{%
\begin{tabular}{lrrrr}
\toprule
Mode & Answer acc. & Schema valid & Exec.\ acc. & Wrong-valid schema \\
\midrule
\mode{prompt\_json} & 19.7 [18.3, 21.1]\% & 61.5 [59.7, 63.3]\% & 12.0 [10.9, 13.2]\% & 49.5 [47.7, 51.2]\% \\
\mode{answer\_only\_schema} & 11.0 [9.9, 12.2]\% & 100.0 [99.9, 100.0]\% & 11.0 [9.9, 12.2]\% & 88.9 [87.8, 90.0]\% \\
\midrule
Delta & -8.7 pts & +38.5 pts & -1.0 pts & +39.4 pts \\
\bottomrule
\end{tabular}%
}
\end{table}

\begin{figure}[t]
\centering
\includegraphics[width=\linewidth]{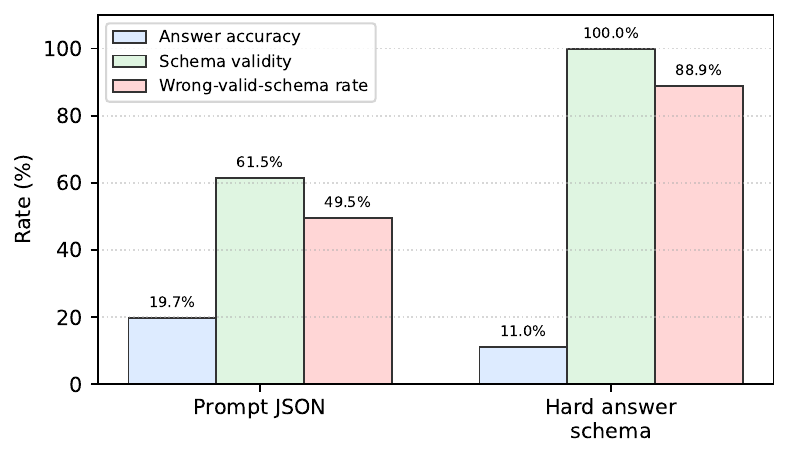}
\caption{Main GPU experiment. Hard answer-only schema decoding reaches 100\% validity, but answer accuracy falls and the wrong-valid-schema rate rises sharply.}
\label{fig:maintradeoff}
\end{figure}

Table~\ref{tab:main} and Figure~\ref{fig:maintradeoff} show the basic tradeoff. Hard answer-only constrained decoding fixes format: validity rises by 38.5 points, from 61.5\% to 100.0\%. It also makes the answers worse: answer accuracy falls by 8.7 points, from 19.7\% to 11.0\%. Executable accuracy nearly ties because the prompt-only mode loses many otherwise correct answers to malformed or schema-incompatible JSON\@. The important shift is in the error type. Wrong-valid-schema outputs rise from 49.5\% to 88.9\%.

That is the systems risk. The parser sees a cleaner stream; the executor sees more well-formed objects carrying wrong content.

\begin{table*}[t]
\centering
\caption{Per-task-family metrics for the main deterministic suite, aggregated across the main-suite sub-3B checkpoints. Each row uses 600 generations per mode. The \mode{tool\_call\_argument} row here is the answer-wrapper version from the main suite, not the executable calendar-object analogue in Table~\ref{tab:toolcall}.}
\label{tab:taskfamilymetrics}
\resizebox{\textwidth}{!}{%
\begin{tabular}{lrrrrrrrrr}
\toprule
Family & Prompt ans. & Prompt valid & Prompt exec. & Schema ans. & Schema valid & Schema exec. & Answer tax & Exec.\ tax & Wrong-valid delta \\
 & \multicolumn{3}{c}{\mode{prompt\_json}} & \multicolumn{3}{c}{\mode{answer\_only\_schema}} & (pts) & (pts) & (pts) \\
\midrule
\mode{arithmetic\_two\_step} & 32.2\% & 33.0\% & 3.3\% & 5.3\% & 100.0\% & 5.3\% & 26.8 & 0.0 & +65.0 \\
\mode{symbolic\_string} & 0.5\% & 79.0\% & 0.5\% & 0.0\% & 99.8\% & 0.0\% & 0.5 & 0.5 & +21.3 \\
\mode{object\_tracking} & 29.0\% & 98.7\% & 29.0\% & 16.5\% & 100.0\% & 16.5\% & 12.5 & 12.5 & +13.8 \\
\mode{boolean\_logic} & 36.8\% & 33.3\% & 27.2\% & 33.3\% & 100.0\% & 33.3\% & 3.5 & 0.0 & +60.5 \\
\mode{tool\_call\_argument} & 0.0\% & 63.3\% & 0.0\% & 0.0\% & 100.0\% & 0.0\% & 0.0 & 0.0 & +36.7 \\
\bottomrule
\end{tabular}%
}
\end{table*}

Table~\ref{tab:taskfamilymetrics} breaks the aggregate apart. Arithmetic and object tracking account for the clearest answer-accuracy losses. Symbolic strings are near floor in both interfaces. Boolean logic is the counterpattern: validity improves without an executable-accuracy loss because prompt-only JSON often contains the right value in a malformed object. The main-suite tool-call row is only an answer-wrapper task, so the executable calendar-object experiment is reported separately.

\subsection{Constraint Tax by Model and Task Suite}

\begin{table*}[t]
\centering
\caption{Constraint tax by model and evaluated task suite. Tax is the clipped accuracy loss of the hard constrained interface relative to the corresponding prompt-only baseline. The main deterministic suite rows remain model-level aggregates; task-family aggregates appear in Table~\ref{tab:taskfamilymetrics}.}
\label{tab:taxbymodeltask}
\resizebox{\textwidth}{!}{%
\begin{tabular}{llrrrrl}
\toprule
Model & Task suite & Answer tax & Exec.\ tax & Validity delta & Wrong-valid delta & Reading \\
 & & (pts) & (pts) & (pts) & (pts) & \\
\midrule
Qwen2.5-0.5B & Deterministic suite & 11.9 & 6.1 & +45.7 & +51.8 & Direct schema hurts accuracy \\
Qwen2.5-1.5B & Deterministic suite & 20.0 & 12.6 & +16.6 & +29.2 & Largest main-suite tax \\
SmolLM2-1.7B & Deterministic suite & 0.0 & 0.0 & +53.2 & +37.4 & Constraint is an accuracy gain \\
Qwen2.5-1.5B & Calendar tool call & not applicable & 43.5 [36.0, 51.0] & +0.0 & +43.5 [35.5, 51.0] & Valid object, wrong arguments \\
Qwen2.5-3B & Boundary deterministic suite & 15.3 [11.2, 19.2] & 15.3 [11.4, 19.3] & +16.3 [13.9, 18.7] & +31.6 [27.7, 35.2] & Tax persists near 3B \\
\bottomrule
\end{tabular}%
}
\end{table*}

Table~\ref{tab:taxbymodeltask} uses the paper's sign convention: positive values are losses induced by the hard constrained interface. The deterministic-suite wrong-valid deltas come from per-instance error-taxonomy labels in the main result JSONL, and the Qwen2.5-3B executable tax is reported because executable accuracy was logged for both boundary modes. The two Qwen sub-3B models pay a direct-schema tax on the deterministic suite. SmolLM2 is the counterexample: its prompt-only JSON baseline is weak enough that hard constraints improve accuracy, so the clipped tax is zero. The calendar tool-call row is the cleanest executable task result because both directly generated modes are 100.0\% schema-valid; the 43.5 point loss is semantic rather than parsing-related. The derived delayed-packaging calendar control has 0.0 executable tax because it preserves the first-stage prompt-only answer and only validates/re-serializes the object.

\subsection{Executable Tool Calls: Valid Objects Can Encode Wrong Decisions}

\begin{table}[t]
\centering
\caption{Calendar tool-call analogue with Qwen2.5-1.5B-Instruct on a commodity GPU\@. Both directly generated modes are 100\% schema-valid, so the hard-schema loss is semantic. The delayed-packaging row is a deterministic re-serialization of the prompt-json first-stage records. Brackets denote 95\% bootstrap confidence intervals.}
\label{tab:toolcall}
\resizebox{\linewidth}{!}{%
\begin{tabular}{lrrrr}
\toprule
Mode & Exec.\ acc. & Schema valid & Wrong-valid schema & Latency \\
\midrule
\mode{tool\_call\_prompt\_json} & 91.5\% [87.5, 95.0] & 100.0\% & 8.5\% [5.0, 12.5] & 1.63s \\
\mode{tool\_call\_delayed\_packaging} & 91.5\% [87.5, 95.0] & 100.0\% & 8.5\% [5.0, 12.5] & 1.63s + pkg. \\
\mode{tool\_call\_schema} & 48.0\% [41.0, 55.0] & 100.0\% & 52.0\% [45.0, 58.5] & 1.69s \\
\midrule
Schema - prompt & -43.5 pts [-51.0, -36.0] & +0.0 pts & +43.5 pts [35.5, 51.0] & +0.06s \\
\bottomrule
\end{tabular}%
}
\end{table}

\begin{figure}[t]
\centering
\includegraphics[width=\linewidth]{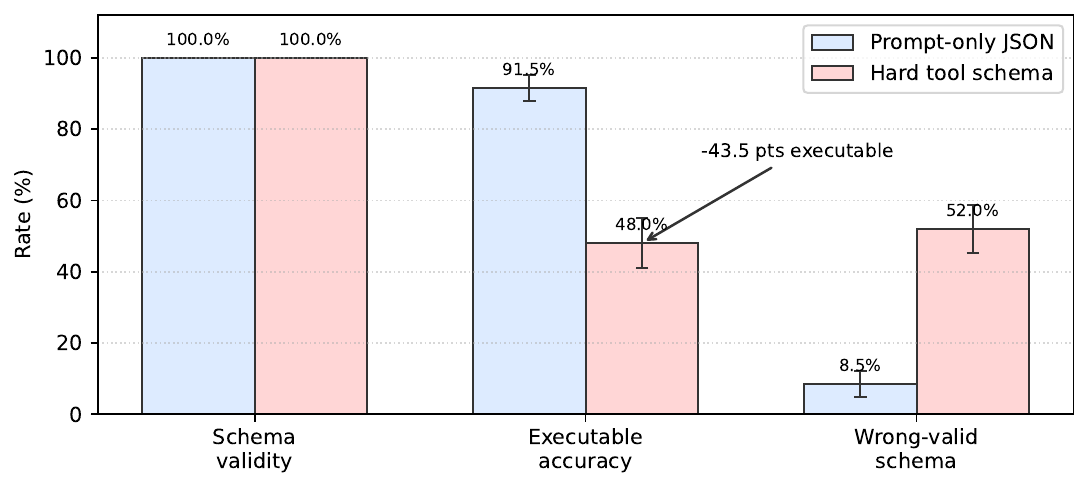}
\caption{Main industry result. Prompt-only JSON and hard tool schema are both 100\% valid, but hard schema decoding loses 43.5 points of executable accuracy and increases wrong-valid-schema failures by 43.5 points.}
\label{fig:toolcallbars}
\end{figure}

Table~\ref{tab:toolcall} and Figure~\ref{fig:toolcallbars} are the most production-like result. The required calendar object is identical in the two directly generated modes. Prompt-only JSON is already 100.0\% schema-valid and reaches 91.5\% executable accuracy. A deterministic delayed-packaging control keeps that semantic result while re-serializing the answer into the executable object. Hard tool-call schema decoding is also 100.0\% valid, but executable accuracy falls to 48.0\%. The paired executable-accuracy delta is -43.5 points, with a 95\% interval of [-51.0, -36.0].

One failure makes the point. The request asks for a 30 minute meeting with Leo. The constrained decoder emits a valid calendar object with the correct date, time, attendee, and topic, but with \mode{duration\_minutes: 180}. A calendar API would accept the object and schedule the wrong meeting.

\begin{table}[t]
\centering
\caption{Primary calendar failure taxonomy counts. Classes sum to 200 per mode; multi-field failures are not double-counted in the primary taxonomy. Delayed packaging is derived from prompt-json first-stage records.}
\label{tab:calendarfailure}
\resizebox{\linewidth}{!}{%
\begin{tabular}{lrrrr}
\toprule
Mode & Correct & Wrong duration & Wrong topic & Multi-field \\
\midrule
\mode{tool\_call\_prompt\_json} & 183 & 8 & 7 & 2 \\
\mode{tool\_call\_delayed\_packaging} & 183 & 8 & 7 & 2 \\
\mode{tool\_call\_schema} & 96 & 102 & 0 & 2 \\
\bottomrule
\end{tabular}%
}
\end{table}

Table~\ref{tab:calendarfailure} localizes the semantic regression: 102 of 104 hard-schema failures are single-field duration errors. In field-occurrence counts, every hard-schema failure includes a wrong \mode{duration\_minutes} value, while only two also include a wrong topic.

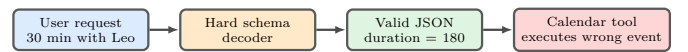
\begin{figure}[t]
\centering
\resizebox{\linewidth}{!}{%
\begin{tikzpicture}[
 font=\scriptsize,
 node distance=0.55cm,
 box/.style={draw=black!65, very thick, rounded corners=3pt, align=center, inner sep=4pt, minimum width=2.5cm, minimum height=0.75cm},
 arrow/.style={-{Latex[length=2mm]}, very thick, draw=black!70}
]
\node[box, fill=ctblue] (request) {User request\\30 min with Leo};
\node[box, fill=ctorange, right=of request] (decoder) {Hard schema\\decoder};
\node[box, fill=ctgreen, right=of decoder] (json) {Valid JSON\\duration = 180};
\node[box, fill=ctred, right=of json] (tool) {Calendar tool\\executes wrong event};
\draw[arrow] (request) -- (decoder);
\draw[arrow] (decoder) -- (json);
\draw[arrow] (json) -- (tool);
\end{tikzpicture}%
}
\caption{Wrong-valid-schema failure path. The object is executable, but the decision encoded inside it is wrong.}
\label{fig:wrongvalid}
\end{figure}

Figure~\ref{fig:wrongvalid} shows the failure path: the parser has nothing to reject, but the tool receives the wrong arguments.

\subsection{Backend Replication and Serving-Stack Effects}

\begin{table}[t]
\centering
\caption{Backend replication for \mode{answer\_only\_schema}. Qwen results replicate across vLLM and SGLang. SmolLM2 is backend-sensitive.}
\label{tab:backend}
\resizebox{\linewidth}{!}{%
\begin{tabular}{lrrl}
\toprule
Model & vLLM acc. & SGLang acc. & Note \\
\midrule
Qwen2.5-0.5B & 2.7\% & 2.7\% & Replicates \\
Qwen2.5-1.5B & 11.7\% & 11.7\% & Replicates \\
SmolLM2-1.7B & 18.7\% & 25.7\% & Backend-sensitive \\
\bottomrule
\end{tabular}%
}
\end{table}

Table~\ref{tab:backend} shows that the Qwen result is backend-stable across vLLM and SGLang for \mode{answer\_only\_schema}. SmolLM2 differs by backend. For MLSys-style reporting, that distinction matters: constrained decoding can depend on serving-engine details, tokenizer handling, grammar implementation, and kernel maturity.

The SGLang replication also required a compatibility patch routing RMSNorm through SGLang's native implementation rather than the current FlashInfer/CUTLASS path. That patch is part of the result: structured-output evaluation on commodity GPUs depends on serving-stack maturity.

\subsection{Direct-Schema Tax at the 3B Boundary}

\begin{table}[t]
\centering
\caption{Qwen2.5-3B-Instruct boundary check on a commodity GPU\@. Direct answer-only schema improves validity but loses answer accuracy and increases wrong-valid-schema outputs.}
\label{tab:3b}
\resizebox{\linewidth}{!}{%
\begin{tabular}{lrrr}
\toprule
Mode & Answer acc. & Schema valid & Wrong-valid schema \\
\midrule
\mode{prompt\_json} & 39.0\% [36.2, 42.0] & 82.6\% [80.2, 85.0] & 43.6\% \\
\mode{answer\_only\_schema} & 23.7\% [21.1, 26.3] & 98.9\% [98.2, 99.5] & 75.2\% \\
\midrule
Delta & -15.3 pts [-19.2, -11.2] & +16.3 pts [13.9, 18.7] & +31.6 pts [27.7, 35.2] \\
\bottomrule
\end{tabular}%
}
\end{table}

Table~\ref{tab:3b} shows that the direct answer-only schema tax does not disappear at the 3B boundary. Qwen2.5-3B improves prompt-only JSON accuracy compared with smaller models, but hard answer schema decoding still loses 15.3 answer-accuracy points while increasing wrong-valid-schema outputs by 31.6 points. The evidence does not support the claim that this issue is limited to the smallest models.

\subsection{Reason Free, Constrain Late}

\begin{table}[t]
\centering
\caption{Local expanded-interface study over four public small models. Delayed constraints and rationale-bearing schemas preserve more semantic accuracy than direct answer-only schemas.}
\label{tab:mlx}
\resizebox{\linewidth}{!}{%
\begin{tabular}{lrrr}
\toprule
Mode & Answer acc. & Schema valid & Exec.\ acc. \\
\midrule
\mode{prompt\_json} & 31.5\% & 70.0\% & 24.5\% \\
\mode{answer\_only\_schema} & 26.8\% & 99.0\% & 26.8\% \\
\mode{rationale\_answer\_schema} & 36.5\% & 97.8\% & 36.5\% \\
\mode{delayed\_constraint} & 40.7\% & 100.0\% & 40.7\% \\
\bottomrule
\end{tabular}%
}
\end{table}

\begin{figure}[t]
\centering
\rotatebox{-90}{\includegraphics[height=\linewidth]{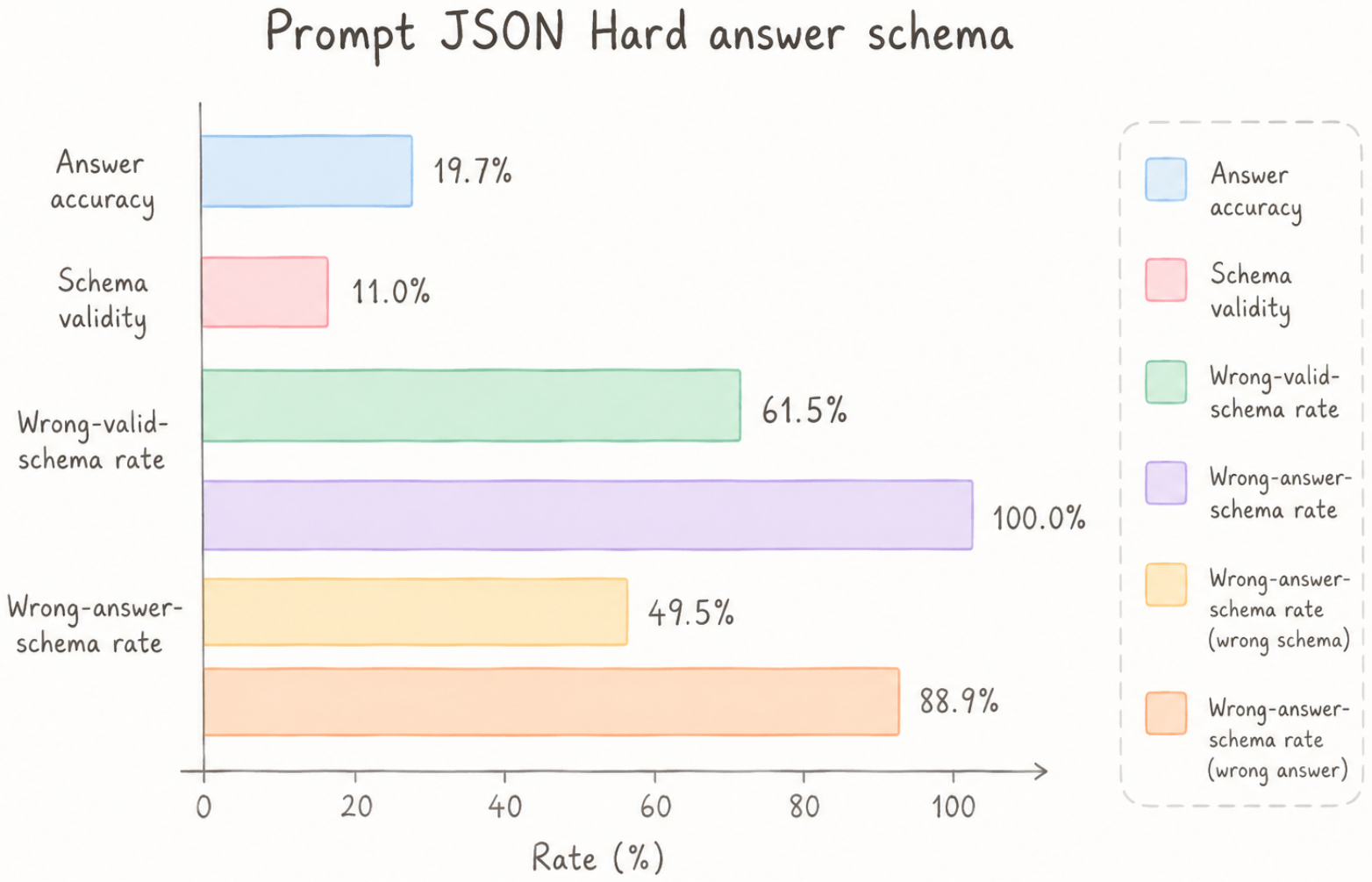}}
\caption{Expanded-interface study. Delayed constraint reaches 100\% validity while preserving the highest executable accuracy among the reported modes.}
\label{fig:mlx}
\end{figure}

The expanded-interface study is smaller than the main GPU run, but it helps identify where the interference enters. Table~\ref{tab:mlx} and Figure~\ref{fig:mlx} show that \mode{answer\_only\_schema} improves validity over \mode{prompt\_json}, but it is not the best executable interface. Short rationales and post-solution packaging do better. \mode{delayed\_constraint} reaches 40.7\% answer and executable accuracy at 100.0\% schema validity.

The engineering pattern is to constrain as late as the product contract allows: let the model solve the task, then project the result into the executable object.

\section{Discussion}

\subsection{Parseability vs. Execution Correctness}

Parseability is a transport property, not a task-success metric. In the tool-call analogue, the directly generated prompt-JSON and hard-schema modes are both 100.0\% valid. The hard-schema mode still loses 43.5 points of executable accuracy. A production system that reports only schema validity would miss the regression.

A structured-output dashboard needs at least four numbers: schema validity, answer accuracy, executable accuracy, and wrong-valid-schema rate. The last one matters most for operations because it counts failures that pass parsers and validators.

\subsection{When Constraints Help Versus Hurt}

The result is not ``schemas are bad''. SmolLM2 improves under direct schema decoding in the main run because its prompt-only JSON baseline is weak. The expanded-interface study also shows gains from rationale-bearing schemas and delayed constraints. The operational boundary is narrower: constraints help when they remove syntax failures without disrupting the task search; they hurt when they convert visible format failures into valid wrong decisions.

\subsection{Serving Engine Effects}

Constrained decoding is also a serving-stack result. Qwen results replicate across vLLM and SGLang, but SmolLM2 is backend-sensitive. The SGLang compatibility patch reinforces the point: the constrained decoder is part of the model interface. Benchmark reports should pin the checkpoint, tokenizer, decoding engine, schema, hardware, and relevant serving-engine patches.

\textbf{Negative results.} Not every constraint behaved like a tax. SmolLM2 improved under direct answer-only schema decoding in the main run because its prompt-only JSON baseline was weak. The expanded-interface study found that rationale-bearing schemas and delayed constraints improved executable accuracy. Some constrained-serving attempts also failed on the tested commodity-GPU path until the SGLang RMSNorm compatibility patch above, and SmolLM2 changed accuracy across vLLM and SGLang. These cases are why the paper treats constraints as a serving-path property, not a model-only property.

\subsection{What the Result Establishes}

For small models, an output schema is part of inference. If it turns invalid outputs into correct executable objects, it is a reliability gain. If it turns visible failures into valid wrong objects, it is a reliability regression. The benchmark keeps those cases separate instead of folding them into a single parse-success number.

\subsection{Deployment Recommendations}

For small-model production systems, the policy implication is concrete.

\begin{enumerate}
 \item Measure prompt-only JSON before hard-constraining the same object.
 \item Track wrong-valid-schema rate as a first-class reliability metric.
 \item Prefer final-answer or delayed constraints when reasoning is needed.
 \item Replicate constrained modes across the intended serving backend.
 \item Treat schema validity as an interface SLO, not as a task-success metric.
\end{enumerate}

\section{Related Work}

Chain-of-thought prompting and GSM8K-style verifier work established two assumptions that this benchmark relies on: intermediate reasoning can matter, and deterministic or executable checks are preferable to judging only surface form~\cite{wei2022cot,cobbe2021gsm8k}. Rather than improving prompts, this paper holds problem instances fixed and changes the output interface.

ReAct and Toolformer showed that model outputs can be used as actions and tool calls~\cite{yao2023react,schick2023toolformer}. The calendar experiment fixes the tool and required argument object, then compares prompt-only JSON against hard schema decoding for the same call. This isolates a lower-level deployment issue: enforcing known tool arguments can still change executable correctness.

Outlines, LMQL, vLLM structured outputs, and SGLang provide practical mechanisms for regex constraints, grammars, JSON schemas, and structured language-model programs~\cite{willard2023outlines,lmql,kwon2023vllm,vllm,zheng2024sglang,sglang}. We treat those mechanisms as part of the measured serving path rather than neutral formatting utilities; the vLLM/SGLang replication is included because constrained decoding behavior is not only a property of the weights.

JSON Schema standardizes structured validation~\cite{jsonschema}. This work keeps that contract but separates validation from task success: a valid object is still checked against the answer normalizer, executor, or trace checker.

\section{Generalization and Reproducibility}

The deterministic setup is a strength for measurement and a limit for generalization. The synthetic families are stress tests for reasoning under output control, not a replacement for broad user workloads. The calendar task is closer to a production tool-call path, but it is still a controlled analogue rather than logged agent traffic.

The expanded-interface study uses fewer examples per task family than the main comparison. Its rationale-schema and delayed-constraint results are design evidence, not a claim that those interfaces always fix the tax. Delayed constraints should be replicated on the serving engines used in deployment before being treated as the default.

The study also does not establish a scaling law. The Qwen2.5-3B boundary still pays a direct-schema tax, but larger models and additional model families are needed to determine when, or whether, this effect disappears. Model coverage is intentionally limited to low-compute, locally accessible, or commodity-GPU-feasible checkpoints.

Structured-output implementations change quickly. Reproducible reports should pin the checkpoint, tokenizer, decoding engine, schema, hardware, and relevant serving patches. The backend-sensitive SmolLM2 result is another example of the same point: constrained decoding belongs to the full serving path, not only to the model weights.

\section{Conclusion}

Structured outputs are an interface contract, but for small language models they are also a decoding intervention. Constraint tax measures the semantic accuracy lost when schemas, regexes, and typed outputs are applied during generation. In the main GPU aggregate, hard schemas improve syntax while reducing answer accuracy. In the calendar tool-call analogue, hard schema decoding keeps validity at 100.0\% but loses 43.5 points of executable accuracy.

For small-model production systems, the evaluation should be paired and interface-aware: report validity and correctness separately, track wrong-valid-schema rate, and test whether the model should solve before the answer is constrained into an executable object.

\appendix
\begingroup
\footnotesize
\section{Calendar Tool-Call Specification}

This appendix gives the exact interface contract used for the calendar tool-call analogue. Each generated instance fills the bracketed slots with deterministic values.

\subsection{Prompt Template}

\begin{verbatim}
You are a calendar assistant. Return only JSON.
Schedule a {duration_minutes} minute meeting
with {attendee} about {topic} on {date} at {start_time}.
Use tool name "create_calendar_event".
\end{verbatim}

For the inspected duration failure discussed in the main text, the natural-language request was:

\begin{verbatim}
Schedule a 30 minute meeting with Leo.
\end{verbatim}

\subsection{Required JSON Schema}

\begin{verbatim}
{
  "type": "object",
  "additionalProperties": false,
  "required": ["tool", "arguments"],
  "properties": {
    "tool": {"const": "create_calendar_event"},
    "arguments": {
      "type": "object",
      "additionalProperties": false,
      "required": [
        "title", "date", "start_time", "duration_minutes",
        "attendee", "topic"
      ],
      "properties": {
        "title": {"type": "string"},
        "date": {
          "type": "string",
          "pattern": "^\\d{4}-\\d{2}-\\d{2}$"
        },
        "start_time": {
          "type": "string",
          "pattern": "^\\d{2}:\\d{2}$"
        },
        "duration_minutes": {
          "type": "integer",
          "minimum": 1
        },
        "attendee": {"type": "string"},
        "topic": {"type": "string"}
      }
    }
  }
}
\end{verbatim}

\subsection{Executable Checker}

The executable checker scores the semantic arguments used by the downstream calendar action. The \mode{title} field is required for schema validity but is not treated as an independent semantic target.

\begin{verbatim}
def normalize(x):
    return str(x).strip().lower()

def calendar_exec_ok(obj, expected):
    if obj["tool"] != "create_calendar_event":
        return False
    args = obj["arguments"]
    return (
        normalize(args["date"]) == expected["date"] and
        normalize(args["start_time"]) ==
        expected["start_time"] and
        int(args["duration_minutes"]) ==
        expected["duration_minutes"] and
        normalize(args["attendee"]) ==
        expected["attendee"] and
        normalize(args["topic"]) == expected["topic"]
    )
\end{verbatim}

\endgroup
\begingroup
\footnotesize
\bibliographystyle{plain}

\endgroup

\end{document}